\documentclass[10pt,twocolumn,letterpaper]{article}

\usepackage[pagenumbers]{iccv} 

\definecolor{iccvblue}{rgb}{0.21,0.49,0.74}
\usepackage[pagebackref,breaklinks,colorlinks,allcolors=iccvblue]{hyperref}

\usepackage{xcolor} 
\definecolor{darkgreen}{RGB}{0,160,0}
\definecolor{orangered}{RGB}{255,69,0} 
\usepackage{booktabs}
\usepackage{multirow}
\usepackage{siunitx}
\usepackage{graphicx} 
\usepackage{overpic}  

\newcommand\blfootnote[1]{%
  \begingroup
  \renewcommand\thefootnote{}\footnote{#1}%
  \addtocounter{footnote}{-1}%
  \endgroup
}

\def\paperID{5128} 
\def\confName{ICCV}
\def\confYear{2025}

\title{HERO: Human Reaction Generation from Videos}

\author{Chengjun Yu$^{1}$, Wei Zhai$^{1,\dagger}$, Yuhang Yang$^{1}$, Yang Cao$^{1,2}$, Zheng-Jun Zha$^{1}$\\
{$^{1}$~University of Science and Technology of China} \\
{$^{2}$~Institute of Artificial Intelligence, Hefei Comprehensive National Science Center}\\
\small{\texttt{\{yucj@mail., wzhai056@, yyuhang@mail., forrest@, zhazj@\}ustc.edu.cn}}
}

\begin{document}

\twocolumn[{%
         \renewcommand\twocolumn[1][]{#1}%
         \maketitle
         \begin{center}
            \centering
            \vspace{-20pt}
            \includegraphics[width=0.98\textwidth]{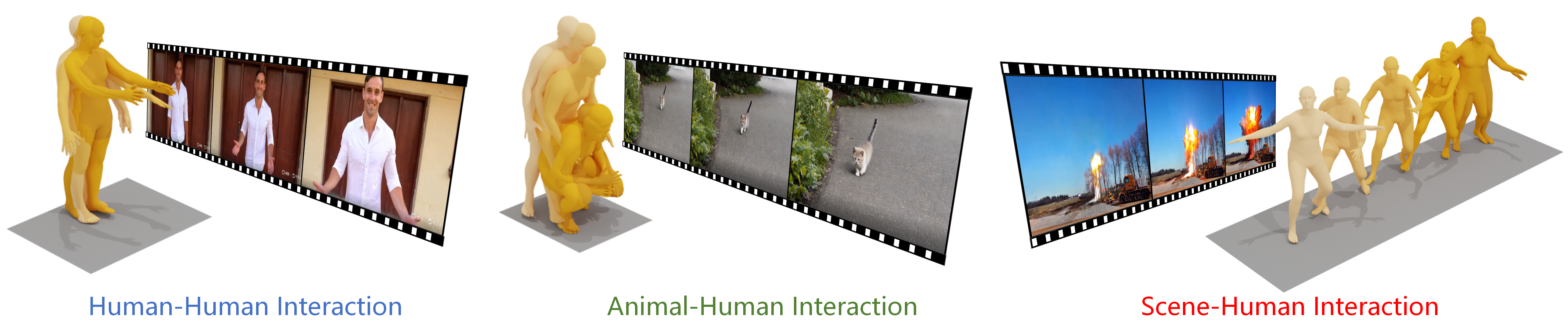}
            \captionof{figure}{\textbf{Overview of our work.} We propose to generate 3D human reactions from RGB videos. To tackle this task, a simple yet powerful framework, HERO, is presented. Furthermore, to facilitate research in this area, we introduce the ViMo dataset, which features a wide range of interaction categories covering three broad ones: \textcolor{blue}{human-human interactions}, \textcolor{darkgreen}{animal-human interactions}, and \textcolor{red}{scene-human interactions}. For the human reactions visualized in the figure, the darker colors indicate the later in time.}
            \label{Fig:teaser}
         \end{center}
}]

\begin{abstract}
\blfootnote{$\dagger$Corresponding Author.} 
Human reaction generation represents a significant research domain for interactive AI, as humans constantly interact with their surroundings. Previous works focus mainly on synthesizing the reactive motion given a human motion sequence. This paradigm limits interaction categories to human-human interactions and ignores emotions that may influence reaction generation. In this work, we propose to generate 3D human reactions from RGB videos, which involves a wider range of interaction categories and naturally provides information about expressions that may reflect the subject's emotions. To cope with this task, we present \textbf{HERO}, a simple yet powerful framework for \textbf{H}uman r\textbf{E}action gene\textbf{R}ation from vide\textbf{O}s. HERO considers both global and frame-level local representations of the video to extract the interaction intention, and then uses the extracted interaction intention to guide the synthesis of the reaction. Besides, local visual representations are continuously injected into the model to maximize the exploitation of the dynamic properties inherent in videos. Furthermore‌, the \textbf{ViMo} dataset containing paired \textbf{Vi}deo-\textbf{Mo}tion data is collected to support the task. In addition to human-human interactions, these video-motion pairs also cover animal-human interactions and scene-human interactions. Extensive experiments demonstrate the superiority of our methodology. The code and dataset will be publicly available at \href{https://jackyu6.github.io/HERO}{https://jackyu6.github.io/HERO}.
\end{abstract}
\section{Introduction}
\label{sec:intro}
Human reaction generation refers to the technology that allows computers to synthesize realistic and natural human reactions based on input signals. It has a wide range of applications in interactive virtual and augmented reality, animation, games, human-robot interaction, and embodied AI, which promises to bring non-player characters (NPCs) to life and enable humanoid robots to understand and respond to social cues and environmental factors in a human-like manner \cite{xu2024regennet, ghosh2024remos, liu2024physreaction}. Previous works \cite{chopin2023interaction, xu2024regennet, liu2023interactive, ghosh2024remos,  liu2024physreaction} focus mainly on synthesizing the motions of reactors conditioned on the motions of actors. These motions are well-organized structured data that computers can easily process. However, intelligent systems such as humanoid robots often use raw visual signals as input. Therefore, human reaction generation from videos is worth exploring.

We note that using human motions as input introduces two limitations in the context of human reaction generation: restricted interaction categories and the absence of emotional information. On the one hand, it limits the interaction types to human-human interactions, overlooking other forms of interaction such as animal-human and scene-human interactions, thereby constraining the machine's capacity to engage with the broader world. On the other hand, the motion data used in prior works do not reflect the emotional states of the characters. In fact, during human-human interactions, the actor's emotions may influence the reactor's decision-making process, particularly when the actor's motion cannot clearly convey interaction intention. Consider the following interaction: ``\textit{the actor walking towards the reactor}''. If the actor approaches the reactor with a neutral facial expression, the reactor may simply remain stationary, awaiting further actions from the actor. When the actor walks towards the reactor in an angry manner, the reactor may tend to respond by stepping back or moving aside. In contrast, if the actor approaches the reactor happily with a smile on their face, the reactor is more likely to move forward and may even initiate interactions such as handshakes or waves. The aforementioned example illustrates that the same action, when associated with different emotional information, could elicit distinct human reactions. Compared to human motion data, RGB videos can encompass a broader range of interaction categories, and naturally contain both the subjects' actions and facial expressions that may reflect emotions (provided the face is visible), without the need for expensive specialized equipment for capture.

In light of the significance to the application and the field discussed above, we propose to generate 3D human reactions from RGB videos. Unlike the previous task on human reaction generation \cite{chopin2023interaction, xu2024regennet, liu2023interactive, ghosh2024remos,  liu2024physreaction} that requires to align the motion of the actor and the synthesized motion of the reactor in the same coordinate system, our aim is to generate a single reactive motion based on a given video, similar to some mainstream tasks of human motion generation, \eg, text-driven human motion generation (text-to-motion generation) \cite{chen2023executing, guo2022generating, guo2022tm2t, petrovich2022temos, tevet2023human, zhang2024motiondiffuse, zhang2023generating}. Nevertheless, a notable distinction exists between our task and text-to-motion generation, particularly with regard to the utilization of representations of the input signals. Specifically, in the text-to-motion paradigm, once the semantic representation of the input text is obtained, it is typically leveraged without further processing, directly serving as a guidance for motion generation. This implies that the model passively adheres to the textual semantics, with limited active comprehension or refinement of the encoded information. In contrast, our objective is not to have the model passively comply with video semantics to synthesize motion. Instead, we aim to enable the model to proactively extract the interaction intention from the visual representations, which is then exploited to guide the generation of the reaction.

The global representation of a video can, to some extent, reflect interaction intention. However, the interaction intention it represents is not ideal due to being derived from the average pooling of frame-level representations \cite{rasheed2023fine, wang2021actionclip, kim2024leveraging}, which treats each frame equally. Notably, certain key-frames within a video are more revealing of interaction intention than other frames. For instance, the global semantics of a video might be described as: ``\textit{A person walks over, initially with no movement in the upper body, then spreads their arms.}'' The frames associated with ``\textit{spreading the arms}'' reveal interaction intention of ``\textit{hugging}'' more significantly than the preceding frames. 

Motivated by the above observation, we propose a global-local representation cross-attention mechanism that simultaneously leverages global and local visual representations obtained from the video encoder \cite{kim2024leveraging} to extract interaction intention. Specifically, it assigns weights to each local frame-level representation on-the-fly based on the global representation of the input video and then performs weighted fusion of per-frame representations to obtain refined information on interaction intention. Next, we employ intention-conditioned self-attention to empower interaction intention to guide reaction generation. We further adopt motion-frame cross-attention to enable the model to obtain the details of the video content, fully leveraging the inherent dynamic properties of the video. We integrate the aforementioned techniques into a new framework, \textbf{HERO}, for human reaction generation from videos.

In addition, we collect the \textbf{ViMo} dataset, which contains video-motion pairs involving human-human, animal-human, and scene-human interactions, featuring diverse interactions of 32 subcategories. It serves as a test bed for model training and evaluation.

Our main contributions can be summarized as follows: \textbf{1)} We propose a new task aimed at generating 3D human reactions from RGB videos with interactivity. Due to the inherent nature of videos, we break through the limitations of previous works with regard to categories and emotions. \textbf{2)} We present HERO, a simple yet powerful framework that extracts interaction intention from the video, and utilize the interaction intention and the dynamic information of the video to synthesize plausible reactions. \textbf{3)} We construct the ViMo dataset that contains video-motion pairs covering a wide range of interaction categories to support the proposed task setting. Extensive experiments demonstrate the feasibility of our task and the superiority of our methodology.

\section{Related Work}
\label{sec:related}

\textbf{Human Motion Generation.} 
The goal of human motion generation is to synthesize human motion conditioned on various signals such as action labels \cite{xu2023actformer, petrovich2021action, guo2020action2motion, cervantes2022implicit, athanasiou2022teach, lucas2022posegpt, yan2019convolutional}, texts \cite{chen2023executing, guo2022generating, guo2022tm2t, petrovich2022temos, tevet2023human, zhang2024motiondiffuse, zhang2023generating}, music \cite{lee2019dancing, siyao2022bailando, aristidou2022rhythm, gong2023tm2d, zhou2023ude, tseng2023edge, li2022danceformer}, speech \cite{ao2022rhythmic, habibie2022motion, ao2023gesturediffuclip, zhu2023taming} and sparse observations \cite{aliakbarian2022flag, castillo2023bodiffusion, du2023avatars, huang2018deep, jiang2022avatarposer}. Different kinds of generative models are utilized in these works for effective modeling. While some works \cite{xu2023actformer, yan2019convolutional, wang2020learning} adopt GAN \cite{goodfellow2014generative} modeling for their methodologies, others \cite{cervantes2022implicit, guo2020action2motion, petrovich2021action, guo2022generating, petrovich2022temos, athanasiou2022teach} utilize the VAE \cite{kingma2013auto} framework to learn probabilistic mapping between conditions and motions. In addition, flow-based models \cite{aliakbarian2022flag, rezende2015variational} are also employed. The emergence of diffusion \cite{sohl2015deep, ho2020denoising} models and autoregressive models has infused new vitality into the field of human motion generation. Diffusion-based methods \cite{shafir2024human, dabral2023mofusion, kim2023flame, tseng2023edge, yuan2023physdiff, kong2023priority, zhou2024emdm, dai2024motionlcm, huang2024stablemofusion} generate motions through a reverse diffusion process from noise. For autoregressive-based methods \cite{zhong2023attt2m, zhang2023generating, gong2023tm2d, guo2022tm2t, jiang2023motiongpt, zhang2024motiongpt, zou2024parco}, motions are first discretized as tokens through vector quantization \cite{van2017neural} so that the Transformer can model the task as a next token prediction problem. Furthermore, some recent works \cite{guo2024momask, pinyoanuntapong2024mmm, pinyoanuntapong2024controlmm, javed2025intermask} propose to exploit masked motion modeling, which achieves real-time and high-fidelity motion generation thanks to the efficient inference process and powerful bidirectional attention. Therefore, we naturally adopt generative masked motion modeling in this work. However, unlike simply following the control signal to synthesize the motion, our task requires the model to extract the interaction intention from the video to generate a plausible human reaction.

\noindent\textbf{Human Reaction Generation.}
Previous works \cite{chopin2023interaction, xu2024regennet, liu2023interactive, ghosh2024remos,  liu2024physreaction} of human reaction generation mainly focus on human-human interactions, i.e., synthesizing the reactor's motion conditioned on the actor's motion. To achieve this goal, \cite{chopin2023interaction} proposes InterFormer, a Transformer network with temporal motion attention and spatial skeleton attention. ReGenNet \cite{xu2024regennet} and ReMoS \cite{ghosh2024remos} adopt diffusion-based models to generate the full body motion of the reactor. \cite{liu2023interactive} designs a method of reaction synthesis with or without objects utilizing social affordance. PhysReaction \cite{liu2024physreaction} introduces a forward dynamics-guided 4D imitation method to generate physically plausible reactions. However, none of the aforementioned works has considered animal-human and scene-human interactions, as well as emotional information of the actor. In contrast, we propose to generate human reaction from RGB videos, which cover human-human, animal-human, and scene-human interactions. In addition to the actor's movement, the video could also provide information about the actor's facial expression, which naturally reflects their emotion.

\noindent\textbf{Human Interaction Dataset.}
There are many datasets on human interaction, as it is a topic that has long been of interest. Some datasets \cite{liang2024intergen, xu2024inter, guo2022multi, hu2013efficient, yin2023hi4d} provide motion data of human-human interactions. However, they lack facial expression information of participants. In other words, these data cannot reflect the emotions of the actors, as videos could. Other datasets \cite{ryoo2010ut, patron2010high, marszalek2009actions, kay2017kinetics} supply videos of human-human interactions. However, most videos involving human-human interactions in these datasets contain the actor and the reactor simultaneously, i.e., they are not filmed from the perspective of the reactor. There are also datasets \cite{fieraru2020three, liu2019ntu, yun2012two, van2016spatio, van2011umpm} that furnish both motion data and videos of human-human interactions. However, their videos suffer from the same issue mentioned above. \cite{ng2020you2me, khirodkar2023ego, khirodkarharmony4d, ryoo2013first, ryoo2015robot, ko2021air} do provide videos from the reactor's view. However, they cover only a few interaction categories and are limited to human-human interactions. To support our work, we build a new dataset that contains video-motion pairs, covering human-human, animal-human, and scene-human interactions.
\section{Method}

\begin{figure*}[t]
  \centering
    \includegraphics[width=1.00\linewidth]{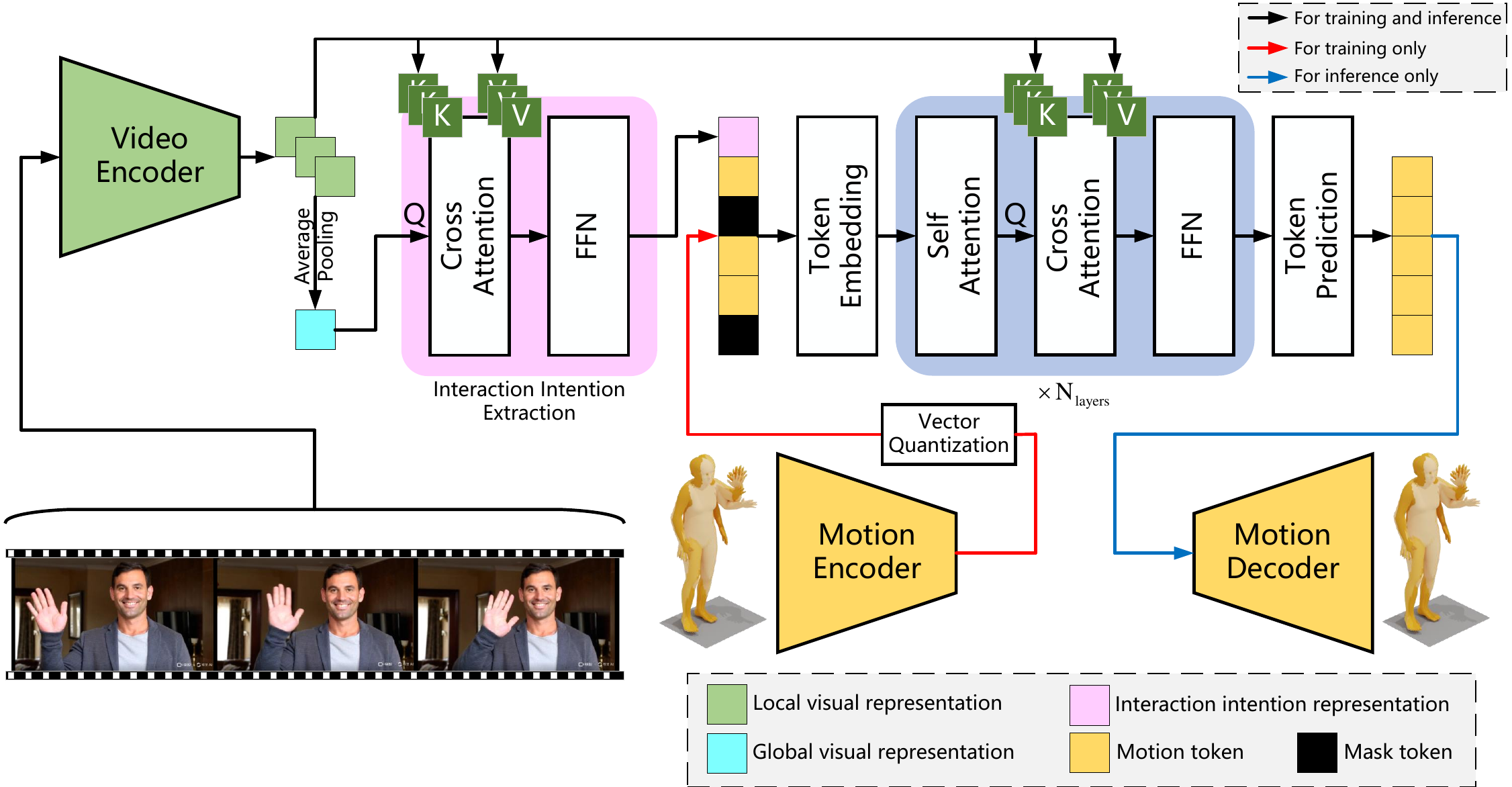}
	\caption{\textbf{The pipeline of HERO.} During training, the video and GT reactive motion are input into HERO. As for inference, only the video is provided. Note that we omit the residual motion refinement (See the end of \cref{sec:3.3}) from the figure for clarity.}
  \label{Fig:pipeline}
\end{figure*}

Given an RGB video $V$ with interactivity, HERO generates a reactive 3D human motion $M$. Specifically, a video $V \in \mathbb{R}^{T \times H \times W \times 3}$ has a spatial resolution of $H \times W$ with $T$ sampled frames, and a motion $M \in \mathbb{R}^{N \times D}$ is a sequence of $N$ poses with $D$ denoting the dimension of pose representations. Our framework consists of three modules: The video encoder (\cref{sec:3.1}) is used to extract visual representations of the input video. The motion VQ-VAE (\cref{sec:3.2}) learns a mapping between the raw motion and discrete code sequences. The reaction generation module (\cref{sec:3.3}) extracts interaction intention from visual representations and uses it to guide the generation of code indices. The pipeline of HERO is illustrated in \cref{Fig:pipeline}.

\subsection{Video Encoder}
\label{sec:3.1}

To extract the representations of the input videos, we employ the video encoder of TC-CLIP \cite{kim2024leveraging}, a paradigm for extending CLIP \cite{radford2021learning} to videos by encoding holistic video information through temporal contextualization. Given a video $V$, the video encoder obtains the per-frame representations $\mathbf{v}_l = [\mathbf{v}_1, \mathbf{ v}_2, ..., \mathbf{v}_T]$, which we call local visual representations. Finally, the global visual representation $\mathbf{v}_g$ is produced by averaging local visual representations formulated as: $\mathbf{v}_g \mathbf{= AvgPool([}\mathbf{v}_1, \mathbf{v}_2, ..., \mathbf{v}_T \mathbf{])}$, where $\mathbf{v}_g \in \mathbb{R}^{d_{vl}}$, $\mathbf{v}_i \in \mathbb{R}^{d_{vl}}$, $i \in \{1, 2, ..., T\}$, $d_{vl}$ is the dimension of the vision-language latent space.

\subsection{Motion VQ-VAE}
\label{sec:3.2}

The functionality of the motion VQ-VAE is to learn discrete representations of motions for subsequent use by the generative module. It transforms the embedding $\mathbf{z}$ output by the motion encoder $\mathbf{E_m}$ into the entries of a learnable codebook $\mathbf{C} = \{\mathbf{c}_k\}_{k=1}^K$ containing $K$ code entries with $\mathbf{c}_k \in \mathbb{R}^{d_c}$, where ${d_c}$ is the dimension of the entries. Specifically, the 1D convolutional motion encoder $\mathbf{E_m}$ first transforms the motion $M \in \mathbb{R}^{N \times D}$ into the latent embedding $\mathbf{z} = \mathbf{E_m}(M) \in \mathbb{R}^{n \times d_c}$ with a temporal downsampling ratio of $\alpha=n/N$. Next, we have the process of vector quantization $\mathbf{Q}$: Each vector in $\mathbf{z}$ is substituted with its closest entry in the codebook $\mathbf{C}$, determined by computing the Euclidean distance. The indices of the sampled entries, i.e., motion tokens $\mathbf{m} \in \mathbb{R}^n$ serve as discrete representations of the input motion. Finally, the motion can be reconstructed by the motion decoder $\mathbf{D_m}$ using the quantized vectors $\mathbf{q=Q(z)} \in \mathbb{R}^{n \times d_c}$ as inputs. Overall, the motion VQ-VAE is trained with the following loss function:
\begin{equation}
\small
\label{Eq:1}
    \mathcal{L}_{vq} = \mathcal{L}_{re} + 
    ||sg[\mathbf{z}]-\mathbf{q}||_{2}^{2} + 
    \beta||\mathbf{z}-sg[\mathbf{q}]||_{2}^{2},
\end{equation}
where $sg[\cdot]$ refers the stop-gradient operator, $\beta$ is a hyper-parameter. For the motion reconstruction, we use $\ell_{1}$ smooth loss: $\mathcal{L}_{re} = \mathcal{L}_{1}^{smooth}(M, \mathbf{D_m}(\mathbf{q})).$
To improve codebook utilization, we apply the exponential moving average and codebook reset following \cite{zhang2023generating, guo2024momask, pinyoanuntapong2024mmm, pinyoanuntapong2024bamm}.

\subsection{Reaction Generation Module}
\label{sec:3.3}

\textbf{Masked Motion Modeling.} 
Recently, generative masked motion modeling \cite{guo2024momask, pinyoanuntapong2024mmm} has achieved high-fidelity and real-time motion generation due to its powerful bidirectional attention and efficient inference process. Therefore, we naturally migrate masked motion modeling to our task. In our experiments, we use tow learnable special tokens: \verb'[PAD]' and \verb'[MASK]'. The \verb'[PAD]' token is utilized to pad shorter motion sequences, enabling the processing of batches containing multiple sequences with different lengths. The \verb'[MASK]' represents input corruption. We first randomly replace a varying fraction of motion tokens $\mathbf{m}$ with the \verb'[MASK]' token to obtain a corrupted sequence $\mathbf{\tilde{m}}$. Then, a masked Transformer model is trained to predict the actual tokens in place of \verb'[MASK]' given conditions $c$ and $\mathbf{\tilde{m}}$. In practice, we linearly project $c$ and $\mathbf{\tilde{m}}$ to the same latent dimension $d_l$. The training objective is to minimize the negative log-likelihood of target predictions:
\begin{equation}
\small
\label{Eq:2}
    \mathcal{L}_{mask} = \mathbb{E}_{\mathbf{m} \in \mathcal{D}}\left[\sum_{\forall\mathbf{\tilde{m}}_i=\verb'[MASK]'} -
    \log p(\mathbf{m}_i|\mathbf{\tilde{m}},c)\right].
\end{equation}

Unlike previous works \cite{guo2024momask, pinyoanuntapong2024mmm} using Transformer encoder, we implement the masked Transformer by stacking $\text{N}_\text{layers}$ Transformer decoder units to better fit our task. We adopt a cosine function to schedule the masking ratio and use the remasking strategy following \cite{guo2024momask}. The inference process begins with an empty canvas, which is represented as a sequence where all tokens are \verb'[MASK]'. Subsequently, the model iteratively predicts more tokens in parallel at each step.

\noindent\textbf{Interaction Intention Extraction (IIE).}
The most different aspect from previous paradigms such as text-to-motion generation is that our goal is not to have the model passively follow control signals, but rather to enable it to proactively extract interaction intention from visual representations to guide reaction generation. In general, the global visual representation $\mathbf{v}_g$ reflects the interaction intention to some extent, as it contains global semantic information about the video. In fact, some key-frames in a video could be more valuable than others for revealing the interaction intention. Therefore, instead of directly using the global visual representation $\mathbf{v}_g$ derived from the average pooling of local visual representations $\mathbf{v}_l$ as the interaction intention representation, we propose the global-local representation cross-attention to extract a refined interaction intention representation. Mathematically, our global-local representation cross-attention is expressed as:
\begin{equation}
\small
\label{Eq:3}
    Att_{gl} = \text{softmax}\left(\frac{\mathbf{Q}_g \mathbf{K}_l^\mathbf{T}}{\sqrt{d_{vl}}}\right)\mathbf{V}_l,
\end{equation}
where the query $\mathbf{Q}_g = \mathbf{v}_g \mathbf{W}_q \in \mathbb{R}^{d_{vl}}$, keys $\mathbf{K}_l = \mathbf{v}_l \mathbf{W}_k \in \mathbb{R}^{T \times d_{vl}}$ and values $\mathbf{V}_l = \mathbf{v}_l \mathbf{W}_v \in \mathbb{R}^{T \times d_{vl}}$ are given by linear projections, $d_{vl}$ is the dimension of the vision-language latent space. Thus, the model can proactively assign weights to each local visual representation on-the-fly based on the global visual representation of the video, and perform a weighted summation of local visual representations to produce an embedding $Att_{gl}$ with refined information on interaction intention. Finally, we input $Att_{gl}$ into an FFN to obtain the interaction intention representation $\mathbf{e}$.

\noindent\textbf{Interaction Intention Injection.}
In order to inject interaction intention into the model so that it can guide the generation of reactions, we adopt intention-conditioned self-attention. Specifically, we first project the interaction intention representation $\mathbf{e}$ and the corrupted motion token sequence $\mathbf{\tilde{m}}$ to the same latent dimension $d_l$, and concatenate them through the temporal dimension. Then we calculate the conditional self-attention between them. In this manner, the interaction intention is effectively integrated into the token sequence through the attention computation process, thereby facilitating the guidance of reaction synthesis by the interaction intention.

\noindent\textbf{Dynamic Information Exploitation (DIE).} Videos inherently possess dynamic attributes that can be well utilized by the model. Intuitively, we linearly project local visual representations to the latent dimension $d_l$, and compute the motion-frame cross-attention to maximize the exploitation of available dynamic information in the video. The calculation of the cross-attention is formulated as follows:
\begin{equation}
\small
\label{Eq:4}
    Att_{mf} = \text{softmax}\left(\frac{\mathbf{Q}_m \mathbf{K}_f^\mathbf{T}}{\sqrt{d_{l}}}\right)\mathbf{V}_f,
\end{equation}
where $\mathbf{Q}_m \in \mathbb{R}^{(n+1) \times d_{l}}$ is derived from linear projection of the output of the previous conditional self-attention, $\mathbf{K}_f \in \mathbb{R}^{T \times d_{l}}$ and $\mathbf{V}_f \in \mathbb{R}^{T \times d_{l}}$ are linear projections of the local visual representations in the latent space. Through this process, the model can take advantage of the inherent dynamic properties of the video to browse the detailed content in the video, and learn the importance of each key-frame in the video for reaction generation. 

\noindent\textbf{Residual Motion Refinement.}
We utilize residual motion refinement as in \cite{guo2024momask, pinyoanuntapong2024bamm} to further improve the quality of the generated motion. In the motion tokenization stage, a residual VQ-VAE (RVQ-VAE) \cite{zeghidour2021soundstream} discretizes the raw motion sequence into multiple code sequences to minimize information loss during the process of quantization. Specifically, each code sequence is produced by a distinct quantization layer, with each quantization layer responsible for encoding the quantization error (namely residual) remaining from the preceding one. In the motion generation stage, the masked Transformer is trained to generate the most informative code sequence (a.k.a. base-layer motion tokens) from the first quantization layer. Using base-layer motion tokens as input, another residual Transformer that has a similar architecture to the masked Transformer predicts the remaining code sequences (a.k.a. residual–layer tokens). Finally, all the code sequences add up to form one sequence used by the decoder for motion reconstruction.

\section{Dataset}
\label{sec:dataset}
\begin{figure}[t]
  \centering
  \small
        \begin{overpic}[width=1.00\linewidth]{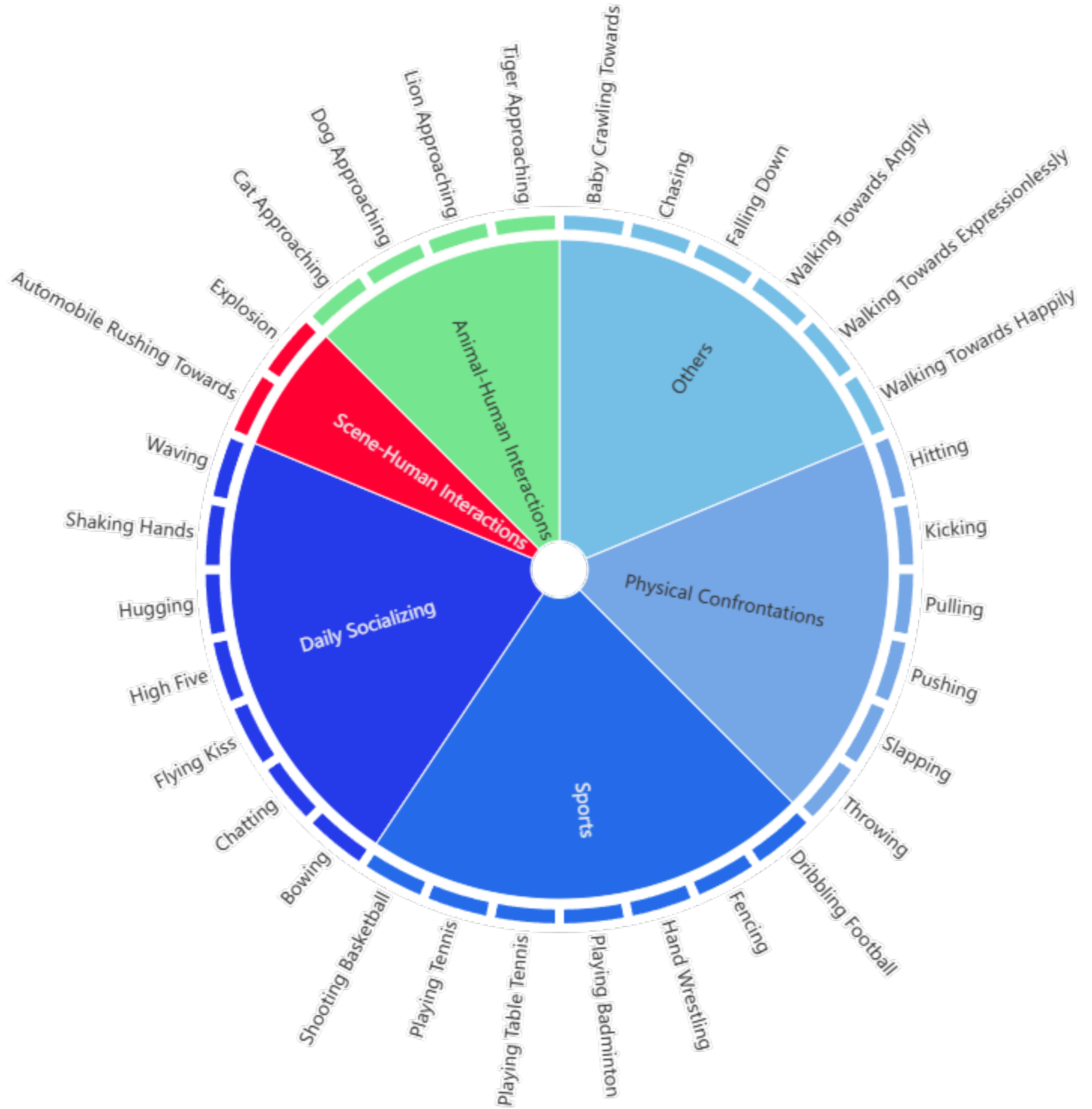}
        \end{overpic}
	\caption{\textbf{ViMo dataset} contains 32 subcategorized interactions, each belonging to one of three broad categories: \textcolor{blue}{human-human interactions}, \textcolor{darkgreen}{animal-human interactions}, and \textcolor{red}{scene-human interactions}. Among them, human-human interactions cover daily socializing, sports, physical confrontations, and others.}
  \label{Fig:dataset}
\end{figure}

We introduce the ViMo dataset, which contains paired video-motion data to support our task setting. \cref{Fig:dataset} provides some information on the categories.

\noindent\textbf{Video Collection.} We mainly acquire RGB videos in two ways: one is to select from publicly available datasets and the other is to utilize video generation models. Specifically, we carefully select 1,335 clips from datasets including UCF101 \cite{soomro2012ucf101}, JPL-Interaction \cite{ryoo2013first}, JPL-Interaction-e \cite{ryoo2015robot}, IDEA400 \cite{lin2023motion}, HMDB51 \cite{kuehne2011hmdb}, Ego-Exo4D \cite{grauman2024ego}, Harmony4D \cite{khirodkarharmony4d}, Ego-Humans \cite{khirodkar2023ego}, etc. The videos we collect adhere to the condition that humans could react to the content if they were seeing it in real life. Long videos are cropped into clips containing only one atomic action. These videos span many categories of human-human interactions, such as daily socializing and sports. However, videos containing certain contents, such as ``\textit{a tiger roaring towards the camera}'' and ``\textit{a car driving fast towards the camera}'' are not easy to find in these datasets. Therefore, we employ several high-performance text-to-video generation models \cite{kling, Vidu, Hailuo, Wanxiang, PixVerse, Jimeng, Mochi} to generate the desired videos and manually filter 1390 high-fidelity ones from them. In addition, to further enrich the diversity and quantity of videos, while obtaining videos of characters with specified actions, we use the following approach: The motions of the actors in Inter-X \cite{xu2024inter} are first rendered as RGB videos from the viewpoint of the reactor. The videos are then converted into a realistic style using Gen-3 Alpha \cite{Gen-3}, a model with the capability of video-to-video generation. Finally, 775 desired videos are selected from them. In summary, applying the methods described above, we collect a total of 3,500 videos featuring humans, animals, and scenes that allow human interaction with the depicted contents. These videos contain 375.61 k frames, with a total duration of 3.83 hours, which cover 32 subcategories. 

\noindent\textbf{Motion Collection and Processing.} The collection of motions is similar to that of videos. On the one hand, we select the corresponding motion data of human reactions from datasets such as Inter-X \cite{xu2024inter}, Harmony4D \cite{khirodkarharmony4d} and Ego-Humans \cite{khirodkar2023ego} based on the collected videos. On the other hand, we employ MoMask \cite{guo2024momask} to generate the desired motion data from text prompts. A manual post-processing step ensues to filter out abnormal motions. We obtain a total of 2,000 motions in a ratio of 3:1 using the above two approaches. The pose representations of all motions are converted to the format used in HumanML3D \cite{guo2022generating}. We scale motions to 20 FPS and crop them into the most semantically informative ones, each with a maximum duration of 10 seconds. For data normalization, motions are then retargeted to a default human skeletal template and properly rotated to face the Z+ direction initially as in \cite{guo2022generating}. As a result, the motions contain a total of 193.61 k frames, equivalent to 2.69 hours, in which the average motion length is 4.84 seconds. 

\noindent\textbf{Video-motion Pairs.} Since all of our motions are hand-picked human reactions corresponding to the content of the collected videos, we can easily pair them up. That is, for each video in our dataset, we can find one or even more corresponding plausible reactions from our motion set. We manually pair each video with a motion to form our base dataset that contains 3,500 video-motion pairs. We also investigate pairing each video with several motions for training in the Sup. Mat. These paired data cover a wide range of interaction categories, with three broad ones: human-human, animal-human, and scene-human interactions. Among them, human-human interactions consist of daily socializing, sports, physical confrontations, and others. More details on ViMo can be found in the Sup. Mat.
\section{Experiments}

\subsection{Experimental Settings}
\label{sec:exp_set}
\noindent\textbf{Dataset Partitioning.} For each subcategory of video-motion pairs, we divide them into the training set and the test set with a ratio of 4:1, resulting in a base dataset (ViMo-base) with 2800 pairs for training and 700 pairs for testing.

\begin{table}[t]
  \centering
  \renewcommand{\arraystretch}{1.}
  \renewcommand{\tabcolsep}{4pt}
  \small
    \begin{tabular}{lccc}
    \toprule
    Method & FID$\downarrow$ & Diversity$\rightarrow$ & MModality$\uparrow$\\ 
    \midrule
    Real & - & $7.954^{\pm0.074}$ & - \\
    \midrule
    MDM \cite{tevet2023human} & $1.688^{\pm0.030}$ & $7.385^{\pm0.088}$ & $\mathbf{2.117^{\pm0.064}}$\\
    MLD \cite{chen2023executing} & $1.565^{\pm0.041}$ & $7.431^{\pm0.090}$ & $2.102^{\pm0.062}$\\
    T2M-GPT \cite{zhang2023generating} & $1.154^{\pm0.038}$ & $7.721^{\pm0.081}$ & $1.936^{\pm0.032}$\\
    BAMM \cite{pinyoanuntapong2024bamm} & $0.930^{\pm0.031}$ & $7.619^{\pm0.055}$ & $1.885^{\pm0.047}$\\
    MoMask \cite{guo2024momask} & $0.856^{\pm0.015}$ & $7.394^{\pm0.056}$ & $1.567^{\pm0.043}$\\
    \textbf{HERO} & $\mathbf{0.427^{\pm0.014}}$ & $\mathbf{7.801^{\pm0.061}}$ & $1.614^{\pm0.040}$\\
    \bottomrule
    \end{tabular}
  \caption{\textbf{Quantitative evaluation on the ViMo test set.} $\pm$ indicates 95\% confidence interval, and $\rightarrow$ means the closer to the real motions the better. \textbf{Bold} face indicates the best result.}
  \label{tab:comparison}
\end{table}

\begin{figure}[t]
\centering
\begin{overpic}[width=0.80\linewidth]{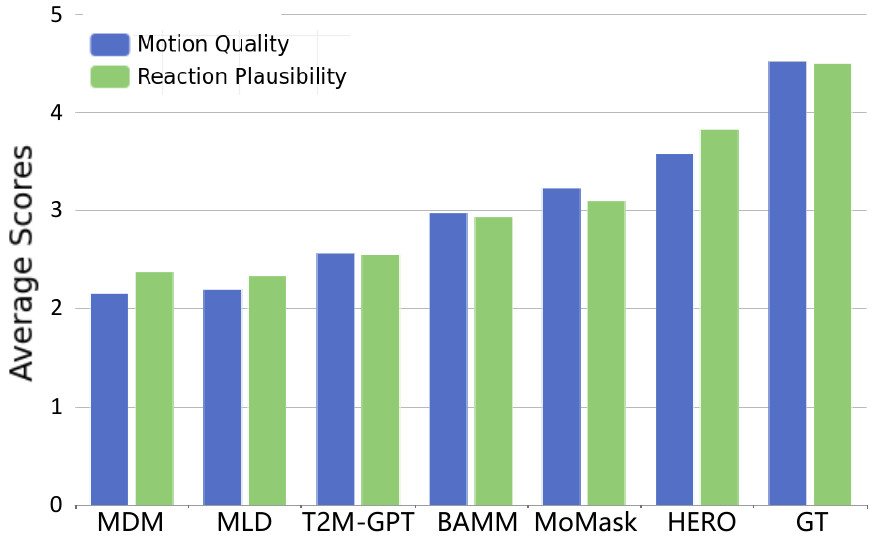}
\end{overpic}
\caption{\textbf{User study results.} The higher the scores, the better.}
\label{Fig:user_study}
\end{figure}

\noindent\textbf{Evaluation Metrics.} We employ metrics that are widely adopted in prior works \cite{guo2020action2motion, guo2022generating, xu2024regennet} on human motion generation, including: Frechet Inception Distance (FID), diversity, and multimodality. 

\noindent\textbf{Implementation Details.} During training and testing HERO, the parameters of the video encoder are frozen. We initialize the RVQ-VAE with the weights obtained by pretraining on HumanML3D \cite{guo2022generating} and then fine-tune it on ViMo for 10 epochs with a batch size of 256. For the Transformers, we train them for 200 epochs with a batch size of 64. Please refer to the Sup. Mat. for more details.

\begin{figure*}[t]
  \centering
    \includegraphics[width=1.00\linewidth]{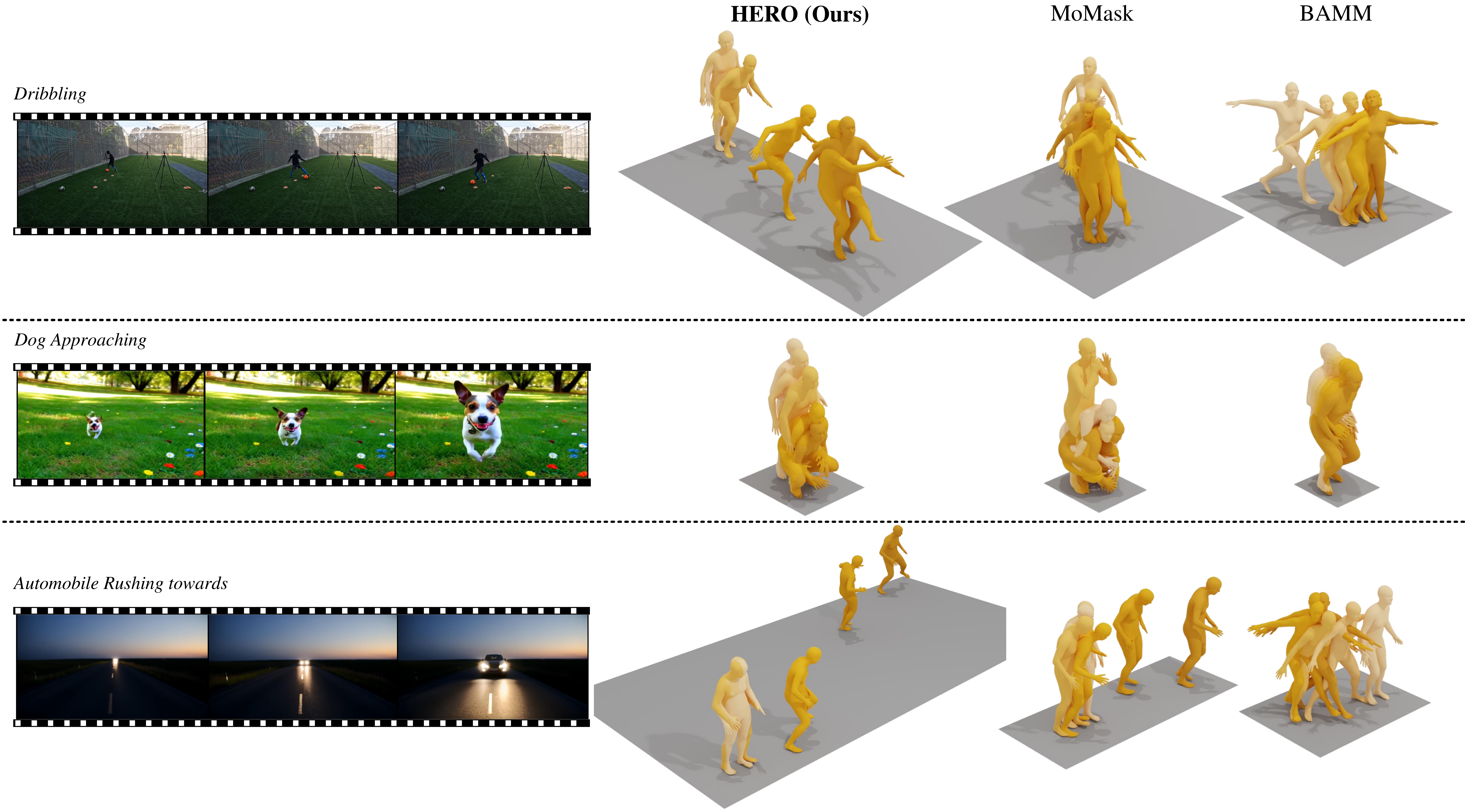}
	\caption{\textbf{Visual comparisons} between the different methods given three distinct videos from ViMo test set.}
  \label{Fig:comparison}
\end{figure*}

\subsection{Comparison to State-of-the-art Approaches}
\label{sec:comparison}
As no prior research has addressed human reaction generation from videos, we reimplement several existing state-of-the-art methods of human-centric generative models as baselines. These methods range from diffusion-based models \cite{tevet2023human, chen2023executing}, autoregressive model \cite{zhang2023generating}, to masked generative models \cite{guo2024momask, pinyoanuntapong2024bamm}. For the sake of fairness, all of their signal encoders are replaced with the video encoder of TC-CLIP \cite{kim2024leveraging}, and all of their VAE/VQ-VAEs are initialized with the weights obtained by pretraining on HumanML3D. 


\begin{table}[t]
  \centering
  \renewcommand{\arraystretch}{1.}
  \renewcommand{\tabcolsep}{6pt}
  \small
    \begin{tabular}{lccc}
    \toprule
    Method & FID$\downarrow$ & Diversity$\rightarrow$ & MModality$\uparrow$\\ 
    \midrule
    Real & - & $7.954^{\pm0.074}$ & - \\
    \midrule
    \textit{w/o} IIE & $0.535^{\pm0.017}$ & $7.671^{\pm0.056}$ & $\mathbf{1.626^{\pm0.042}}$\\
    \textit{w/o} DIE & $0.521^{\pm0.011}$ & $7.769^{\pm0.068}$ & $1.443^{\pm0.052}$\\
    \textbf{HERO} & $\mathbf{0.427^{\pm0.014}}$ & $\mathbf{7.801^{\pm0.061}}$ & $1.614^{\pm0.040}$\\
    \bottomrule
    \end{tabular}
  \caption{\textbf{Quantitative ablation studies.} \textit{w/o} means without. IIE and DIE represent Interaction Intention Extraction and Dynamic Information Exploitation, respectively.}
  \label{tab:ablation}
\end{table}

\begin{figure}[]
\centering
\begin{overpic}[width=0.92\linewidth]{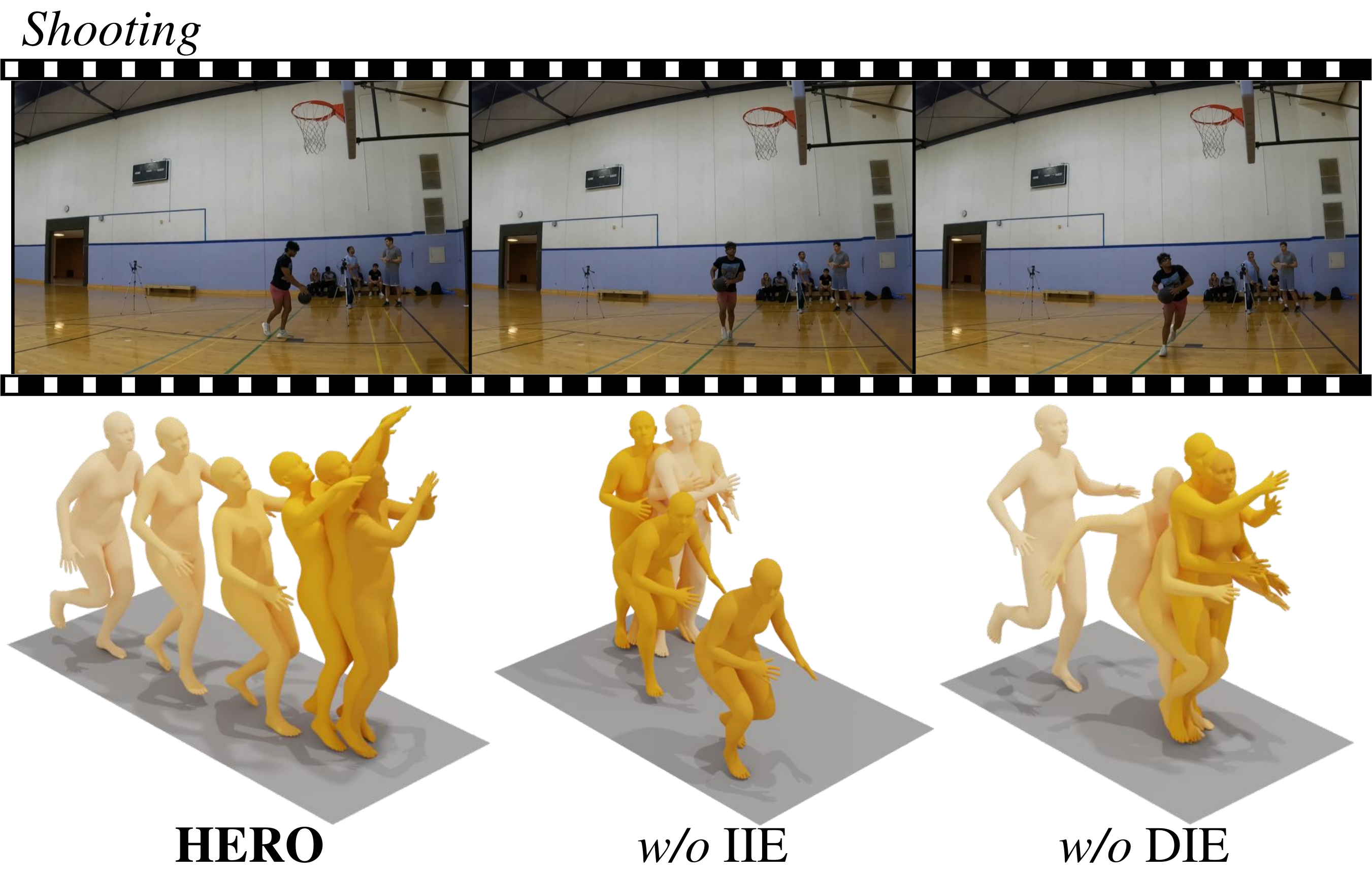}
\end{overpic}
\caption{\textbf{Qualitative ablation studies} given the same video. A plausible reaction would be to defend against the shooting.}
\label{Fig:ablation}
\end{figure}

As reported in \cref{tab:comparison}, HERO outperforms the baselines on FID and diversity, suggesting that it learns the reaction patterns well. Although multimodality is important, \cite{guo2024momask} highlights its role as a secondary measure which should be evaluated alongside primary performance metrics such as FID. The qualitative comparisons in \cref{Fig:comparison} show that HERO is more capable of generating plausible reactions than other methods.

\noindent\textbf{User Study.} We ask participants to rate (1 to 5, the higher the better) the motions they observe in terms of the quality of the reactive motion (motion quality), and the plausibility of the reaction given the video (reaction plausibility) as in \cite{ghosh2024remos}. For each method, 50 reactions are generated using the same video pool from the ViMo test set. We report the average scores from the feedback of 40 participants in \cref{Fig:user_study}, excluding the feedback that did not pass our validation checks. In both metrics surveyed, HERO-generated reactions receive the highest scores, except for GT.

\begin{figure*}[t]
  \centering
    \includegraphics[width=1.00\linewidth]{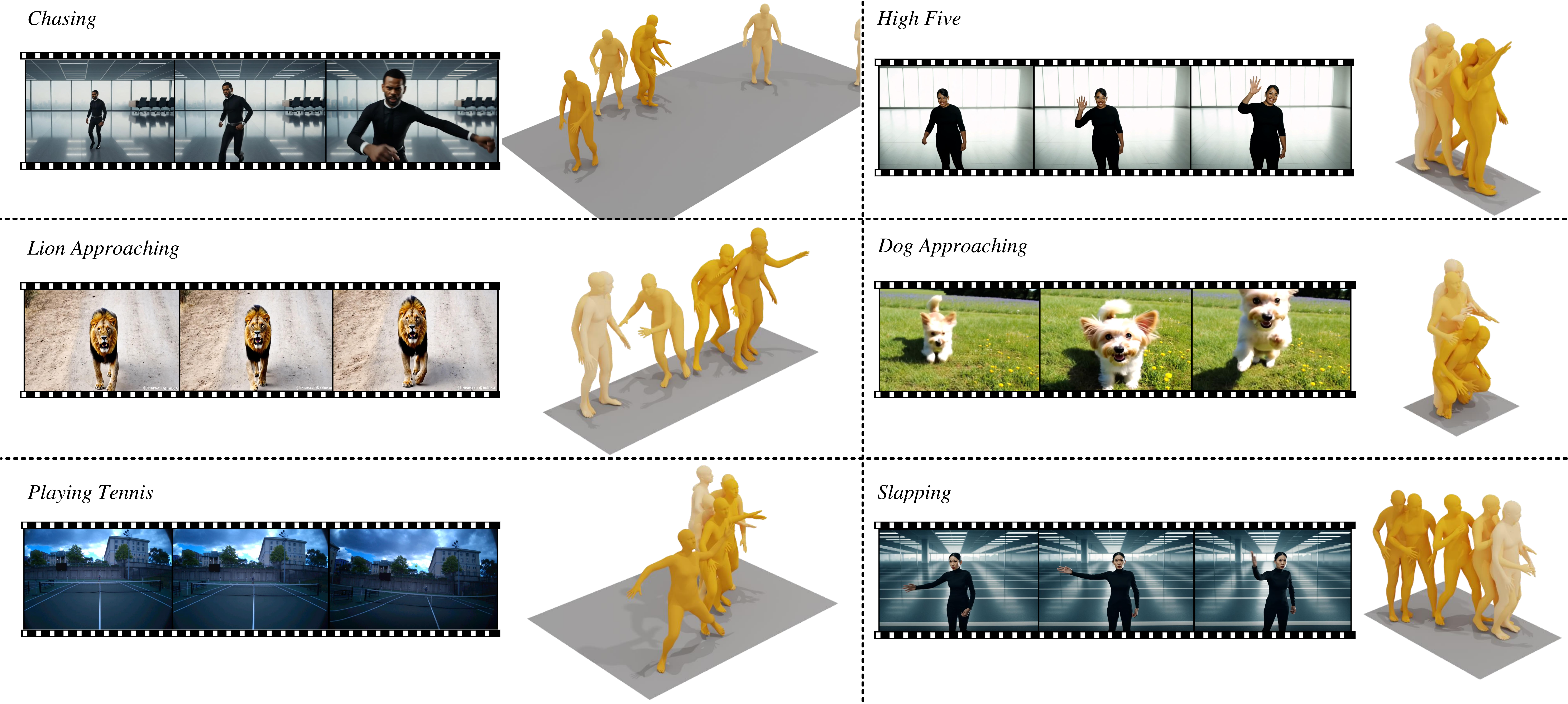}
	\caption{\textbf{Visualized cases of generating on the Unseen set.} Note that the data of the subcategories to which each video in the figure belongs are not utilized when training HERO on the \textbf{Seen} set.}
  \label{Fig:unseen}
\end{figure*}

\subsection{Ablation Study}
We can observe from \cref{tab:ablation} that both interaction intention extraction (IIE) and dynamic information exploitation (DIE) effectively bring the generated reactions closer to the real distribution. \cref{Fig:ablation} illustrates that the absence of IIE may produce an implausible reaction, while the absence of DIE leads to a decrease in motion quality.

\subsection{Performance Analysis}
\label{sec:analysis}

\noindent\textbf{Different Broad Categories of Interactions.} \cref{Fig:teaser} and \cref{Fig:comparison} show the reactions synthesized by HERO in different broad categories of interactions: human-human, animal-human and scene-human interactions. The visualization results validate the effectiveness of HERO in different broad categories of interactions. The quantitative results are presented in the Sup. Mat.

\begin{figure}
\centering
\small
\begin{overpic}[width=0.98\linewidth]{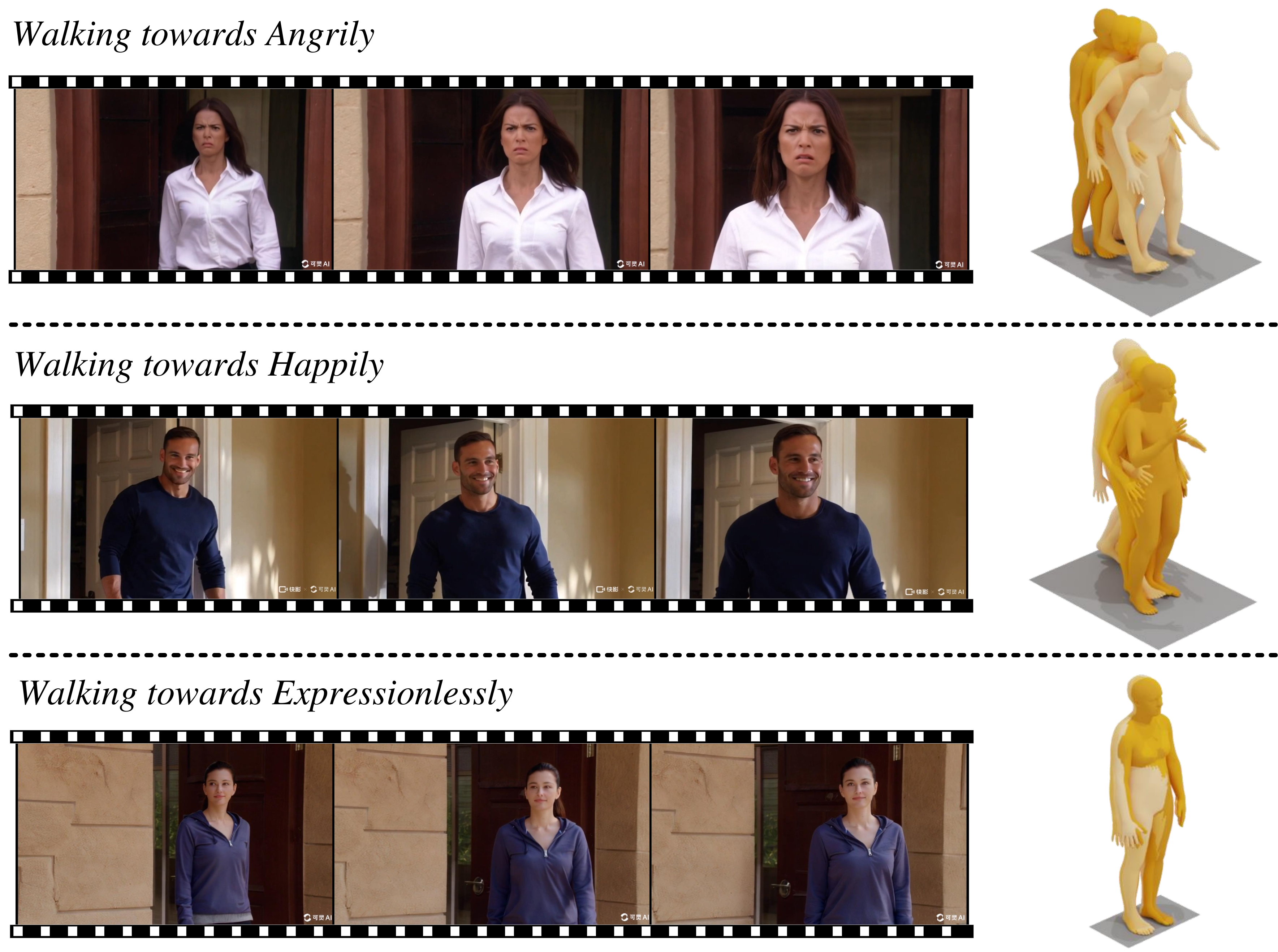}
\end{overpic}
\caption{\textbf{Visualized cases of ``\textit{walking towards}'' with different emotions.} The left side of each row in the figure is the input video and the right side is the reaction generated by HERO.}
\label{Fig:diff_emo}
\end{figure}

\noindent\textbf{The Same Action with Different Emotions.} As demonstrated in \cref{Fig:diff_emo}, HERO can generate different plausible reactions depending on the subjects' different emotions, even if the subjects' actions are the same.

\noindent\textbf{Generalization Ability.} We further divide the 3500 pairs in the base dataset into two parts, \textbf{Seen} and \textbf{Unseen}. The \textbf{Seen} set covers 26 subcategories of interactions, and the \textbf{Unseen} set contains the remaining 6 subcategories. To verify the generalization ability of HERO, we train it on the \textbf{Seen} set and use it to generate reactions on the \textbf{Unseen} set. The visualized cases in \cref{Fig:unseen} show that HERO dose have a certain generalization ability for unseen video categories. More experimental results can be found in the Sup. Mat.

\section{Conclusion}
In this paper, we introduce a new task focused on generating 3D human reactions from RGB videos with interactivity. With the collected ViMo dataset containing video-motion pairs, we train HERO, a model that extracts interaction intention from input video and uses it to guide the generation of the plausible human reaction. Extensive experiments demonstrate the effectiveness and superiority of HERO. We hope that this work can contribute new insights into 3D human motion generation involving interaction.

\noindent\textbf{Limitations and Future Work.} Currently, the categories and volume of video-motion pairs still need to be expanded, and our framework is as simple as possible to verify the feasibility of the task. In the future, we plan to integrate the bidirectional autoregressive approach \cite{pinyoanuntapong2024bamm} into our framework so that the model can predict the length of the reaction by itself. Also, we can incorporate physical constraints \cite{yuan2023physdiff} into the pipeline to eliminate artifacts such as floating and foot sliding, often observed with human motion generation models. In addition, another interesting direction may draw on depth estimation \cite{yang2024depth} and body shape estimation \cite{cai2024smpler} to study how to synthesize a more vivid and reasonable reaction based on proximity and body shape of the subject in the video.

{
    \small
    \bibliographystyle{ieeenat_fullname}
    \bibliography{main}
}

\clearpage
\setcounter{page}{1}
\maketitlesupplementary

\section{More Details on ViMo Dataset}
\label{sec:sup_dataset}
\noindent\textbf{Explanation of Utilizing Synthetic Data.}
As described in \cref{sec:dataset}, since some of the data are difficult to acquire from publicly available datasets, we adopt generative models to obtain some desired data to form part of our dataset. To guarantee the quality of the data, we leverage several cutting-edge generative models \cite{kling, Vidu, Hailuo, Wanxiang, PixVerse, Jimeng, Mochi, guo2024momask} that are known for their superior performance, and manually screen the synthetic data. It is worth mentioning that synthetic data is customizable and low-cost. In addition, the videos generated by the models have a wide variety of characters and backgrounds. Another advantage of using synthetic data is that it does not violate the subject's right to likeness because none of the persons in the synthetic videos are real.

\noindent\textbf{Pose Representation.}
We use the same pose representation as in HumanML3D \cite{guo2022generating}, which is over-parameterized, expressive, and neural network friendly \cite{liang2024intergen}, and has been widely adopted in recent works \cite{tevet2023human, chen2023executing, zhang2023generating, guo2024momask, pinyoanuntapong2024bamm}. Each pose in the motion sequence is defined by $(\dot{r}^a, \dot{r}^x, \dot{r}^z, r^y, j^p, j^v, j^r, c^f)$, where $\dot{r}^a \in \mathbb{R}$ is root angular velocity along the Y-axis; $\dot{r}^x \in \mathbb{R} \text{ and } \dot{r}^z \in \mathbb{R}$ are root linear velocities on the XZ-plane; $r^y \in \mathbb{R}$ is root height; $j^p \in \mathbb{R}^{3j}, j^v \in \mathbb{R}^{3j} \text{ and } j^r \in \mathbb{R}^{6j}$ are the local joints positions, velocities and rotations in root space, with j indicating the number of joints; $c^f \in \mathbb{R}^4$ are binary features representing foot ground contacts derived from thresholding the velocities of the heel and toe joints.

\begin{figure*}[]
  \centering
  \small
        \begin{overpic}[width=1.00\linewidth]{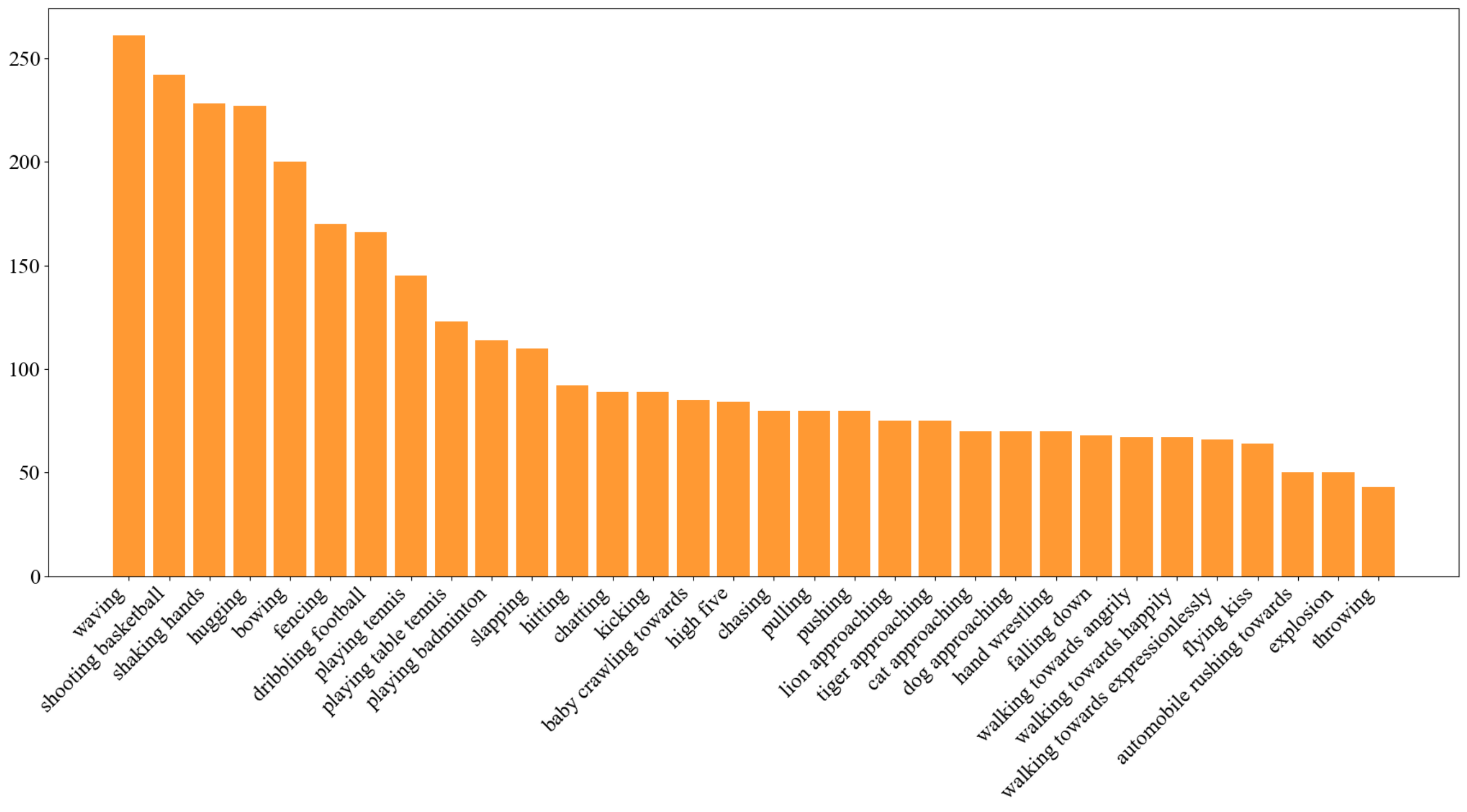}
        \end{overpic}
	\caption{\textbf{Distribution of the video data in ViMo dataset.}}
  \label{Fig:video_cat_dis}
\end{figure*}

\begin{figure*}[]
  \centering
  \small
        \begin{overpic}[width=1.00\linewidth]{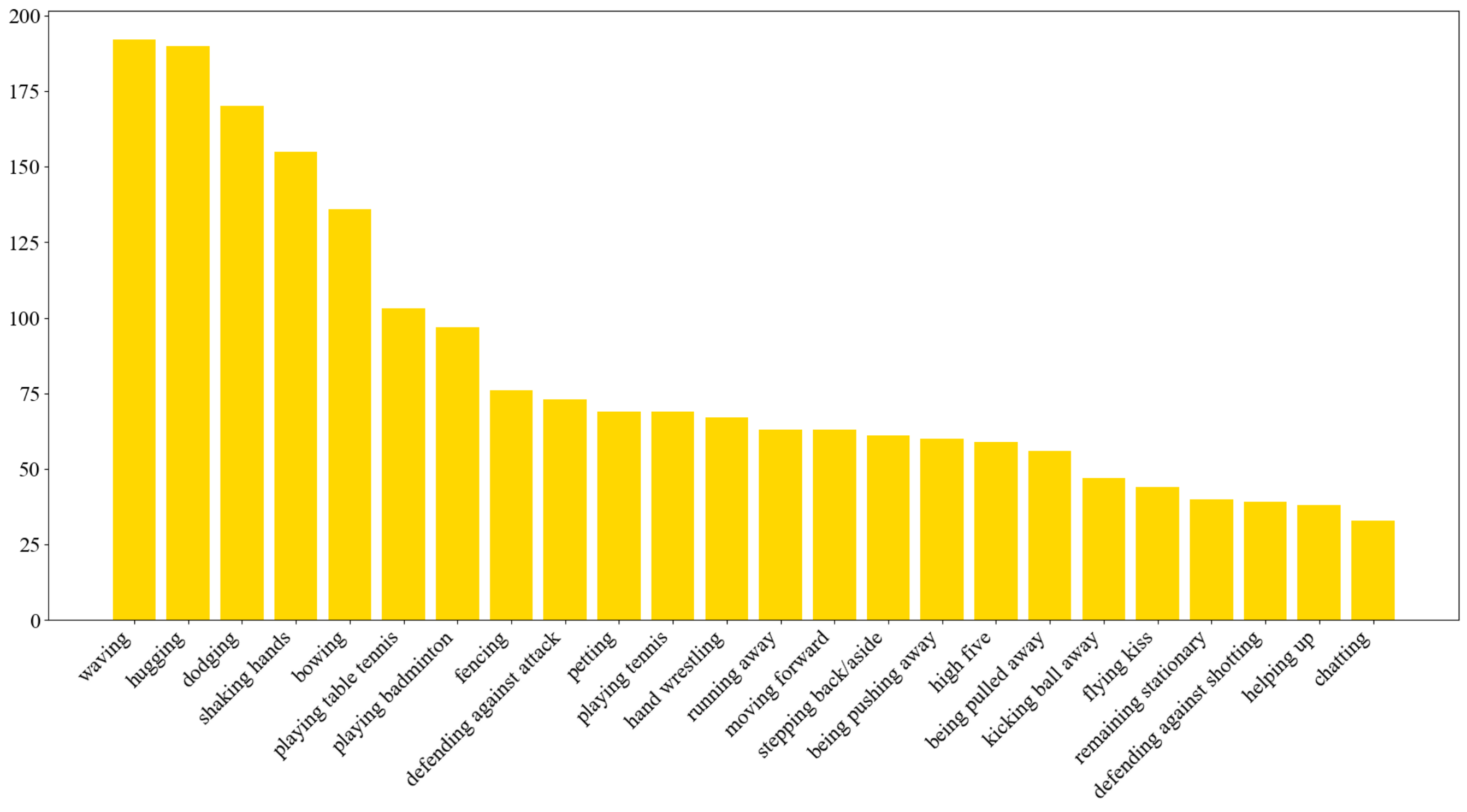}
        \end{overpic}
	\caption{\textbf{Distribution of the motion data in ViMo dataset.}}
  \label{Fig:motion_cat_dis}
\end{figure*}

\noindent\textbf{Data Distribution.} \cref{Fig:video_cat_dis} and \cref{Fig:motion_cat_dis} provide some information on the distribution of the videos and the motion data, respectively.

\section{More Details on Evaluation Metrics}
\label{sec:sup_metrics}
In line with prior practices \cite{guo2022generating, tevet2023human}, each experiment is conducted 20 times, and the reported values of the metrics indicate the mean along with a confidence interval of 95\%.

\noindent\textbf{FID.} Frechet Inception Distance (FID) is adopted as the principal metric to evaluate the overall quality of the generation, which is calculated between the feature distribution of the generated motions and the feature distribution of the real motions. The feature extractor is from \cite{guo2022generating}.

\noindent\textbf{Diversity.} Diversity measures the variability and richness of motions, which is calculated by averaging the Euclidean distances of 300 randomly sampled pairs of motions.

\noindent\textbf{MultiModality.} MultiModality measures the diversity of human motion generated from the same video. Specifically, it represents the average variance for a single video by computing the Euclidean distances of 10 sampled pairs of generated motions. For each video, we generate the motion 30 times.
 
\noindent\textbf{Explanation of the Absence of Action Recognition Accuracy.} \cite{xu2024regennet, liu2023interactive} use a pretrained model to classify the generated motions and calculate the action recognition accuracy. However, this does not apply to our task. In our setup, motions generated from videos in one category may belong to multiple categories, as long as they are plausible reactions to the videos. For example, the reactions to “\textit{walking towards happily}” might be a wave, a handshake, or even a hug. Besides, the reactions to “\textit{hitting}” might be to dodge, parry, or counterattack... Therefore, we do not use action recognition accuracy as one of our evaluation metrics.

We stress that action recognition accuracy is not a primary performance metric as FID is. Some other works \cite{liu2024physreaction, ghosh2024remos} of human reaction generation do not use it either. \cite{xu2024regennet, liu2023interactive} use specific pose representations, they train the classification model to obtain the feature extractor mainly for measuring FID, which is incidentally used to measure accuracy. In contrast, the motions in our ViMo dataset are provided with the same pose representation as in HumanML3D \cite{guo2022generating}. Thus, we can naturally employ the high-quality feature extractor from \cite{guo2022generating}, which is trained on much more motion data than in our dataset. 

In the context of human reaction generation, the accuracy is mainly used to measure plausibility, i.e., whether the generated reaction is plausible for the input. However, as mentioned earlier, the accuracy is not applicable to our task. To compensate for the lack of plausibility evaluation, we investigate reaction plausibility through visualization and user study (\cref{sec:comparison}) following \cite{ghosh2024remos}.

\section{More Details on Implementation}
\label{sec:sup_imp}
Our models are implemented using PyTorch. TC-CLIP \cite{kim2024leveraging} is initialized with the weights of CLIP \cite{radford2021learning} with ViT-B/16 and then pretrain on Kinetics-400 \cite{kay2017kinetics}, a large-scale action recognition dataset with a total of 400 action classes and around 240k training videos. Therefore, the video encoder's action recognition capability is enhanced on the basis of its general recognition ability. This applies to our scenario, as most of the data in our ViMo dataset are about human-human interactions. Following \cite{kim2024leveraging, wang2021actionclip}, 16 frames are sampled from each video before entering the video encoder, and all frames are resized to a uniform resolution of 224 × 224. Our RVQ-VAE has the same architecture as in MoMask \cite{guo2024momask} and the pose representation in ViMo is the same as in HumanML3D \cite{guo2022generating}, so the RVQ-VAE can be initialized with the weights obtained by pretraining on HumanML3D. As for Transformers, we set the number of Transformer decoder units $\text{N}_\text{layers}=6$ in our models. Our models are trained with the AdamW \cite{loshchilov2017decoupled} optimizer. The learning rate reaches 2e-4 after 20 and 250 iterations with a linear warm-up schedule for training RVQ-VAE and Transformers, respectively. All models are trained using 2 NVIDIA A40 GPUs.

\section{More Experimental Results}
\label{sec:sup_exp}
\noindent\textbf{Ablation Study.} We present experimental results with different number of Transformer decoder units $\text{N}_\text{layers}$ in \cref{tab:sup_ablation}. We notice that HERO achieves the best overall performance when setting $\text{N}_\text{layers}=6$.

\begin{table}[t]
  \centering
  \renewcommand{\arraystretch}{1.}
  \renewcommand{\tabcolsep}{7pt}
  \small
    \begin{tabular}{lccc}
    \toprule
    $\text{N}_\text{layers}$ & FID$\downarrow$ & Diversity$\rightarrow$ & MModality$\uparrow$\\ 
    \midrule
    & - & $7.954^{\pm0.074}$ (R.) & - \\
    \cmidrule{1-4}
    4 & $0.659^{\pm0.014}$ & $7.549^{\pm0.051}$ & $1.572^{\pm0.059}$\\
    5 & $0.639^{\pm0.018}$ & $7.604^{\pm0.075}$ & $1.642^{\pm0.046}$\\
    6 & $\mathbf{0.427^{\pm0.014}}$ & $\mathbf{7.801^{\pm0.061}}$ & $1.614^{\pm0.040}$\\
    7 & $0.589^{\pm0.015}$ & $7.621^{\pm0.068}$ & $\mathbf{1.649^{\pm0.051}}$\\
    8 & $0.683^{\pm0.014}$ & $7.499^{\pm0.074}$ & $1.550^{\pm0.041}$\\  
    \bottomrule
    \end{tabular}
  \normalsize
  \caption{\textbf{Ablation studies on the number of Transformer decoder units $\text{N}_\text{layers}$.} $\pm$ indicates 95\% confidence interval, and $\rightarrow$ means the closer to the real motions the better. \textbf{Bold} face indicates the best result. R. means real motions.}
  \label{tab:sup_ablation}
\end{table}

\noindent\textbf{Different Broad Categories of Interactions.} The quantitative results on different broad categories of interactions are reported in \cref{tab:diff_cate}. Numerically, HERO performs well in the broad category of human-human interactions. The suboptimal performance of HERO in other broad categories may be due to limited data in these categories. It is also possible that these broad categories are harder for the model to learn. Nevertheless, from the visualization results shown in \cref{Fig:comparison} of the main text, HERO can generate plausible reactions in all broad categories.

\begin{table}[t]
  \centering
  \renewcommand{\arraystretch}{1.}
  \renewcommand{\tabcolsep}{4pt}
  \small
    \begin{tabular}{lccc}
    \toprule
    Category & FID$\downarrow$ & Diversity$\rightarrow$ & MModality$\uparrow$\\ 
    \midrule
    Scene-Human R.& - & $3.666^{\pm0.111}$ & - \\
    Scene-Human & $3.726^{\pm0.249}$ & $3.676^{\pm0.184}$ & $1.916^{\pm0.063}$\\
    \midrule
    Animal-Human R.& - & $6.755^{\pm0.387}$ & - \\
    Animal-Human & $1.809^{\pm0.075}$ & $6.695^{\pm0.380}$ & $1.362^{\pm0.044}$\\
    \midrule
    Human-Human R.& - & $7.807^{\pm0.063}$ & - \\
    Human-Human & $0.421^{\pm0.011}$ & $7.737^{\pm0.043}$ & $1.454^{\pm0.042}$\\
    \bottomrule
    \end{tabular}
  \caption{\textbf{Evaluation results on different broad categories of interactions.} R. means real motions.}
  \label{tab:diff_cate}
\end{table}

\noindent\textbf{The Same Action with Different Emotions.} We can find in \cref{tab:diff_emo} that the reaction generation of ``\textit{walking towards angrily}'' and ``\textit{walking towards happily}'' is much more difficult to learn than that of ``\textit{walking towards expressionlessly}''. However, \cref{Fig:diff_emo} of the main text shows that HERO is able to synthesize distinct plausible reactions according to different emotions, even if the actions in the videos are the same.

\begin{table}[t]
  \centering
  \renewcommand{\arraystretch}{1.}
  \renewcommand{\tabcolsep}{4pt}
  \small
    \begin{tabular}{lccc}
    \toprule
    Category & FID$\downarrow$ & Diversity$\rightarrow$ & MModality$\uparrow$\\ 
    \midrule
    Angrily R.& - & $3.411^{\pm0.215}$ & - \\
    Angrily & $5.347^{\pm0.430}$ & $4.234^{\pm0.241}$ & $2.014^{\pm0.064}$\\
    \midrule
    Happily R.& - & $5.074^{\pm0.290}$ & - \\
    Happily & $5.215^{\pm0.627}$ & $3.835^{\pm0.216}$ & $1.831^{\pm0.056}$\\
    \midrule
    Expressionlessly R.& - & $1.548^{\pm0.082}$ & - \\
    Expressionlessly & $0.900^{\pm0.146}$ & $1.383^{\pm0.123}$ & $0.589^{\pm0.039}$\\
    \bottomrule
    \end{tabular}
  \caption{\textbf{Evaluation results on ``\textit{walking towards}'' with different emotions.} R. means real motions.}
  \label{tab:diff_emo}
\end{table}

\noindent\textbf{Generalization Ability.} We train the models on the \textbf{Seen} set and evaluate them on the \textbf{Unseen} set. The quantitative comparisons reported in \cref{tab:general} show that HERO achieves the best FID score, and values of diversity and multimodality comparable to those of other methods. Some visualized cases can be found in \cref{Fig:unseen} of the main text.

\begin{table}[t]
  \centering
  \renewcommand{\arraystretch}{1.}
  \renewcommand{\tabcolsep}{4pt}
  \small
    \begin{tabular}{lccc}
    \toprule
    Method & FID$\downarrow$ & Diversity$\rightarrow$ & MModality$\uparrow$\\ 
    \midrule
    Real & - & $7.935^{\pm0.061}$ & - \\
    \midrule
    BAMM \cite{pinyoanuntapong2024bamm} & $2.541^{\pm0.079}$ & $7.143^{\pm0.082}$ & $\mathbf{2.192^{\pm0.062}}$\\
    MoMask \cite{guo2024momask} & $2.477^{\pm0.049}$ & $\mathbf{7.151^{\pm0.084}}$ & $2.009^{\pm0.064}$\\
    \textbf{HERO} & $\mathbf{2.156^{\pm0.054}}$ & $7.095^{\pm0.066}$ & $2.093^{\pm0.057}$\\
    \bottomrule
    \end{tabular}
  \caption{\textbf{Quantitative evaluation on the Unseen set.}}
  \label{tab:general}
\end{table}

\begin{table}[t]
  \centering
  \renewcommand{\arraystretch}{1.}
  \renewcommand{\tabcolsep}{6pt}
  \small
    \begin{tabular}{lccc}
    \toprule
    Num. of T.P. & FID$\downarrow$ & Diversity$\rightarrow$ & MModality$\uparrow$\\ 
    \midrule
    Real & - & $7.954^{\pm0.074}$ & - \\
    \midrule
    2800 & $0.427^{\pm0.014}$ & $7.801^{\pm0.061}$ & $1.614^{\pm0.040}$\\
    5600 & $0.398^{\pm0.012}$ & $7.815^{\pm0.069}$ & $1.529^{\pm0.058}$\\
    8400 & $0.376^{\pm0.014}$ & $7.833^{\pm0.064}$ & $1.422^{\pm0.049}$\\
    \bottomrule
    \end{tabular}
  \caption{\textbf{Evaluation results on different number of training pairs.} T.P. means training pairs.}
  \label{tab:more_pairs}
\end{table}

\noindent\textbf{Training on More Pairs.} In addition to ViMo-base mentioned in \cref{sec:exp_set}, we manually pair each video and two motions in the training set of ViMo-base to form twice as many training data as in the base dataset, that is, 5600 video-motion pairs (ViMo-T2). Similarly, we also obtain 8400 video-motion pairs (ViMo-T3) for training, which is three times the number of training pairs in the base dataset. Note that ViMo-base, ViMo-T2, and ViMo-T3 share the same test set.

As shown in \cref{tab:more_pairs}, training on more data pairs leads to consistent improvements in FID and diversity. The decay in multimodality implies a decrease in the diversity of reactions generated by the model to a single video, but perhaps indirectly indicates that the model is becoming more confident in the generated motions. Although multimodality is undoubtedly important, \cite{guo2024momask} emphasizes its role as a secondary metric that should be evaluated alongside primary performance metrics like FID. Theoretically, we can pair more than 100 k video-motion pairs. The lack of training pairs can be alleviated to some extent by pairing more data.

\section{Discussion}
Another implementation to deal with our task might be to utilize a combination of a multimodal large language model (MLLM) and a text-to-motion generative model. Specifically, the MLLM gives a textual description of the corresponding reaction based on the input video, and the text-to-motion generative model synthesizes the motion based on this description. However, after extensive testing of MLLMs \cite{chen2024motionllm, maaz2024video, lin2024video, jin2024chat}, we observe that despite being able to describe the video content well, these MLLMs struggle to output texts that accurately describe reactive motions in detail. In addition, MLLMs tend to generate redundant information (which reduces the quality of the text used for motion generation), despite being asked not to. These issues prevent them from outputting textual descriptions like those written by humans in HumanML3D, making it difficult for them to perform as an ideal preceding module for the text-to-motion generative model. In contrast, our framework adopts an end-to-end one-stage manner that does not require manually crafted text prompts and is much more efficient.

Nevertheless, an interesting direction is to utilize the multi-turn conversation capability of MLLMs to instruct the model for motion reasoning, generation, and editing \cite{jiang2024motionchain}. In addition to the discussion in the main text, modeling the details of hand motions to generate more expressive reactions is also a future work.

\noindent\textbf{Broader Impacts.} According to \citet{rosenblum2010}, vision is the dominant sense in human perception, accounting for approximately 80\%--90\% of sensory input. Similarly, \citet{jensen1996} estimates that around 83\% of information acquired by humans comes from vision. Compared to texts, videos contain denser information. Our work is no longer limited to enabling machines to passively follow instructions to generate motion. Instead, we aim to empower them to proactively explore how to interact with the world through visual signals. We believe that our work has the potential for a wide range of applications, especially in the emerging field of embodied intelligence.

\end{document}